\documentclass[sigconf]{acmart}

\usepackage{dashrule}
\usepackage{multirow}    % 支持单元格多行合并
\usepackage{booktabs}    % 优雅的表格线 \toprule, \midrule, \bottomrule
\usepackage{array}       % 加强表格列格式
\usepackage{makecell}
\usepackage{tabularx}

\usepackage{url}
\usepackage{color}
\usepackage{xcolor}

\usepackage{amssymb}
\usepackage{amsfonts}
\usepackage{arydshln}
\usepackage{graphicx}
\usepackage{caption}

\setcopyright{none}
\makeatletter
\def\@conference{}
\def\@acmConference@shortname{}
\def\@acmConference@name{}
\def\@acmConference@date{}
\def\@acmConference@venue{}
\def\@acmConference@location{}
\def\@acmISBN{}
\def\@acmDOI{}
\makeatother
\begin{document}

\title{PuzzleBench: A Fully Dynamic Evaluation Framework for Large Multimodal Models on Puzzle Solving}

\author{Zeyu Zhang}
\authornote{Both authors contributed equally to this research.}
\email{zeyuzhang@sjtu.edu.cn}
\author{Zijian Chen}
\authornotemark[1]
\email{zijian.chen@sjtu.edu.cn}
\affiliation{%
  \institution{Shanghai Jiao Tong University}
  \city{Shanghai}
  \country{China}
  \postcode{200240}
}
\author{Zicheng Zhang}
\email{zzc1998@sjtu.edu.cn}
\affiliation{%
  \institution{Shanghai Jiao Tong University}
  \city{Shanghai}
  \country{China}
  \postcode{200240}
}
\author{Yuze Sun}
\email{StevenSun@sjtu.edu.cn}
\affiliation{%
  \institution{Shanghai Jiao Tong University}
  \city{Shanghai}
  \country{China}
  \postcode{200240}
}
\author{Yuan Tian}
\email{ee_tianyuan@sjtu.edu.cn}
\affiliation{%
  \institution{Shanghai Jiao Tong University}
  \city{Shanghai}
  \country{China}
  \postcode{200240}
}
\author{Ziheng Jia}
\email{jzhws1@sjtu.edu.cn}
\affiliation{%
  \institution{Shanghai Jiao Tong University}
  \city{Shanghai}
  \country{China}
  \postcode{200240}
}
\author{Chunyi Li}
\email{lcysyzxdxc@sjtu.edu.cn}
\affiliation{%
  \institution{Shanghai Jiao Tong University}
  \city{Shanghai}
  \country{China}
  \postcode{200240}
}
\author{Xiaohong Liu}
\email{xiaohongliu@sjtu.edu.cn}
\affiliation{%
  \institution{Shanghai Jiao Tong University}
  \city{Shanghai}
  \country{China}
  \postcode{200240}
}
\author{Xiongkuo Min}
\email{minxiongkuo@sjtu.edu.cn}
\affiliation{%
  \institution{Shanghai Jiao Tong University}
  \city{Shanghai}
  \country{China}
  \postcode{200240}
}
\author{Guangtao Zhai}
\email{zhaiguangtao@sjtu.edu.cn}
\affiliation{%
  \institution{Shanghai Jiao Tong University}
  \city{Shanghai}
  \country{China}
  \postcode{200240}
}

\renewcommand{\shortauthors}{Zhang et al.}
\renewcommand\footnotetextcopyrightpermission[1]{}
\settopmatter{printacmref=false} % Removes citation information below abstract
% \author{Anonymous Authors}
%% You do not have to enter your paper ID



% synthesizes visual question-answering (VQA) samples through a modular pipeline: randomly selects visual elements under specific conditions; a visual generation module organizes these elements into structured puzzle images according to the open-ended rules; and a puzzle information extraction module records sampling conditions and open-ended rules to automatically generate corresponding questions and uniquely verifiable answers. 

\begin{abstract}
Large Multimodal Models (LMMs) have demonstrated impressive capabilities across a wide range of multimodal tasks, achieving ever-increasing performance on various evaluation benchmarks. However, existing benchmarks are typically static and often overlap with pre-training datasets, leading to fixed complexity constraints and substantial data contamination issues. Meanwhile, manually annotated datasets are labor-intensive, time-consuming, and subject to human bias and inconsistency, leading to reliability and reproducibility issues. To address these problems, we propose a fully dynamic multimodal evaluation framework, named \textbf{Open-ended Visual Puzzle Generation (OVPG)}, which aims to generate fresh, diverse, and verifiable evaluation data automatically in puzzle-solving tasks.
Specifically, the OVPG pipeline consists of a raw material sampling module, a visual content generation module, and a puzzle rule design module, which ensures that each evaluation instance is primitive, highly randomized, and uniquely solvable, enabling continual adaptation to the evolving capabilities of LMMs. 
Built upon OVPG, we construct \textbf{PuzzleBench}, a dynamic and scalable benchmark comprising 11,840 VQA samples. It features six carefully designed puzzle tasks targeting three core LMM competencies, visual recognition, logical reasoning, and context understanding. PuzzleBench differs from static benchmarks that quickly become outdated. It enables ongoing dataset refreshing through OVPG and a rich set of open-ended puzzle designs, allowing seamless adaptation to the evolving capabilities of LMMs.
\end{abstract}


\begin{teaserfigure}
  \includegraphics[width=\linewidth]{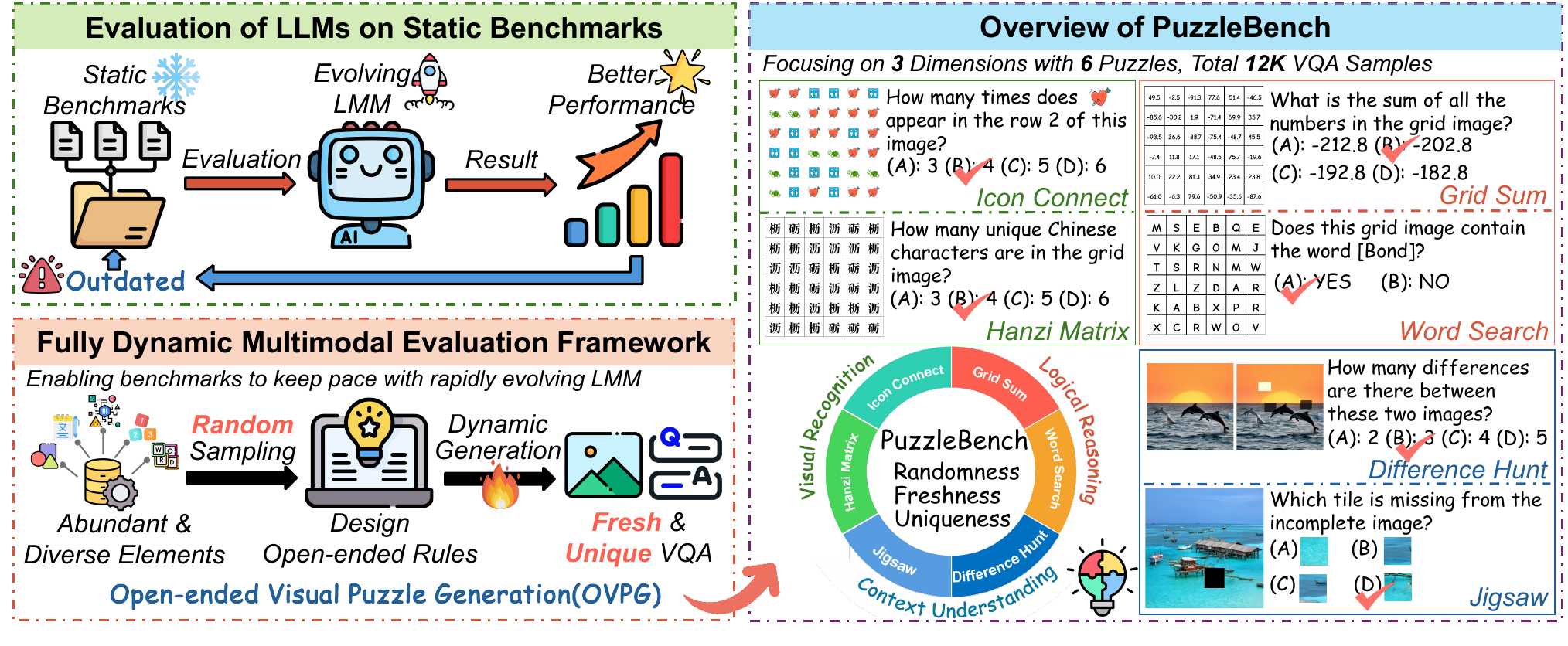}
   %\includegraphics[width=0.8\linewidth]{egfigure.eps}
    %\vspace{-7pt}
   \caption{PuzzleBench is constructed upon a fully dynamic multimodal evaluation framework, Open-ended Visual Puzzle Generation (OVPG), which contains 11,840 VQA samples characterized by high randomness, data freshness, and answer uniqueness. It encompasses six puzzle tasks, targeting three core LMM capabilities, visual recognition, logical reasoning, and context understanding, ensuring the diversity and comprehensiveness of the evaluation. Unlike static benchmarks that soon become outdated, \textbf{PuzzleBench} keeps evolving through continuous dataset \textbf{refreshing} via \textbf{OVPG}.}
   \label{fig:pipeline}
\end{teaserfigure}

\maketitle
\acmConference{}{}{}

% \centering
% \includegraphics[width=1.0\linewidth]{Figs/teaser.pdf}
% \caption{XXX}
% \label{fig:teaser}

% \begin{figure*}[t]
%   \centering
%   \includegraphics[width=0.98\linewidth]{Figs/teaser.pdf}
%     \vspace{-7pt}
%    \caption{PuzzleBench is constructed upon a fully dynamic multimodal evaluation framework, Visual Puzzle Bootstrapping (VPB), which contains 11,840 VQA samples characterized by high randomness, data freshness, and answer uniqueness. It encompasses six puzzle tasks, targeting three core LMM capabilities, i.e., visual recognition, logical reasoning, and in-context understanding, thus ensuring the diversity and comprehensiveness of the evaluation. In contrast to static benchmarks, PuzzleBench facilitates continuous dataset \textbf{refreshing} through VPB, enabling it to keep pace with the rapid development in LMMs.}
%    \label{fig:pipeline}
%    \vspace{-0.25cm}
% \end{figure*}

\section{Introduction}

\begin{figure*}[h]
  \centering
  \includegraphics[width=1\linewidth]{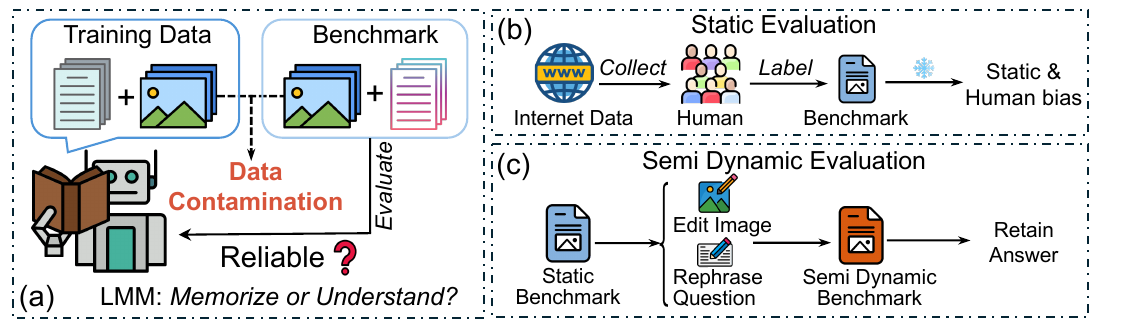}
   \caption{(a) Data contamination: existing benchmarks often overlap with training data, compromising evaluation reliability. (b) Static and manually annotated benchmarks: fixed data distributions and human bias limit adaptability and scalability. (c) Semi-dynamic approaches: editing existing data offers limited novelty due to the need to preserve original answer consistency.}
   \label{fig2}
\end{figure*}

 Large Multimodal Models (LMMs) \cite{wang2024qwen2,reid2024gemini,GPT-4o,wu2024deepseek,chen2024expanding} have recently achieved remarkable success across various multimodal tasks, particularly excelling in visual recognition \cite{huang2024relationvlm,wang2024cogvlm,chen2024gaia}, logical reasoning \cite{xiao2024logicvista,wu2024logical,wang2024exploring}, and context understanding \cite{zhang2024internlm,zhou2024visual}. These advancements are closely connected to the development of diverse benchmarks across different domains and dimensions, which play a critical role in understanding the capabilities and limitations of LMMs, thus promoting their comprehensive development.

While the evaluation of LMMs has received increasing attention, it still faces three major unresolved issues. As shown in Fig. \ref{fig2}:
(1) \textit{Data contamination} \cite{deng2023investigating,golchin2023data,xu2024benchmark} poses a serious threat to evaluation reliability. Since LMMs are typically trained on large-scale web data, there is often significant overlap between evaluation benchmarks and pre-training datasets. This overlap makes it difficult to determine whether a model has genuinely acquired a capability or is simply memorizing seen inputs.  (2) \textit{Static benchmarks} lack adaptability. Once a benchmark is constructed, its content and distribution remain fixed, which limits its effectiveness in evaluating rapidly evolving LMMs. They inherently lack scalability and flexibility, making them prone to obsolescence as model capabilities advance. (3) \textit{Manual annotation} remains the dominant approach for benchmark construction. This process is time-consuming, labor-intensive, and prone to human bias and inconsistency, thereby compromising the objectivity and scalability of evaluations. Effective evaluation of LMMs requires benchmarks that are dynamic, automated, and capable of continuous refreshing; however, constructing such benchmark poses significant challenges.

Existing dynamic evaluation approaches primarily concentrate on single text modalities. For instance, Fan \textit{et al.} \cite{fan2024nphardeval} developed NPHardEval, a benchmark with a monthly dynamic update mechanism specifically targeting reasoning abilities of LLMs. Similarly, DyVal \cite{zhu2023dyval} introduced a dynamic data generation method informed by graph structures, providing controllable complexity to mitigate data contamination. More recently, Yang \textit{et al.} \cite{yang2024dynamic} proposed a vision-language bootstrapping approach capable of dynamically generating visual question-answering (VQA) samples by editing images or rephrasing questions. Nonetheless, this method merely modifies existing data without refreshing the underlying data source, a limitation we categorize as \textit{semi-dynamic evaluation}. Furthermore, alterations through such bootstrapping approaches may not consistently preserve the original intent and corresponding ground-truth answers, even when auxiliary validation agents are employed, thereby failing to achieve complete dynamism. An effective benchmark must be dynamic, automated, and refreshable. Building such a system for multimodal evaluation remains highly non-trivial.

In this work, we introduce \textbf{Open-ended Visual Puzzle Generation (OVPG)}, a fully dynamic multimodal evaluation framework ensuring \textbf{random} image generation process, \textbf{fresh} data source and \textbf{unique} ground-truth answers. OVPG comprises three key modules: the raw material sampling module, the puzzle rule design module, and the visual content generation module. The sampling module randomly selects raw material from diverse source pools, ensuring randomness. The rule design module defines the puzzle target and specifies task-specific arrangement rules for the sampled raw material. The visual content generation module arranges sampled raw material into puzzle images based on puzzle rules, ensuring structural validity and high visual freshness.

Using \textbf{OVPG}, we dynamically created \textbf{PuzzleBench}, generating 11,840 unique VQA samples across six specifically designed puzzle tasks, targeting the three core cognitive dimensions of LMMs: \textit{visual recognition, logical reasoning, and context understanding.} We evaluated multiple state-of-the-art LMMs, including closed-source APIs such as GPT-4o \cite{GPT-4o} and gemini-1.5 \cite{reid2024gemini}, as well as 11 popular open-source LMMs like Qwen series \cite{wang2024qwen2,bai2025qwen25vl}, internVL2.5 series \cite{chen2024expanding} and Deepseek-VL2 \cite{wu2024deepseek}. Our experiments demonstrated that \textbf{OVPG} significantly mitigates data contamination issues with complete dynamism. Furthermore, our comprehensive evaluation revealed that current LMMs still exhibit considerable room for improvement, highlighting clear directions for future model development.

In summary, our contributions can be summarized as three-fold:
\begin{itemize}
\item[(1)] Introducing the fully dynamic multimodal evaluation framework \textbf{OVPG}, which ensures randomness in the generation process, freshness of the generated data, and uniqueness of questions and ground-truth answers, thereby significantly reducing data contamination.
\item[(2)] Dynamically generating \textbf{PuzzleBench}, a benchmark composed of 11,840 unique VQA samples spanning six puzzle tasks. These tasks are constructed to evaluate LMMs across three core cognitive dimensions, ensuring both diversity in task formats and depth in capability assessment.
\item[(3)] Conducting comprehensive evaluations on popular LMMs reveals that these models still struggle with fine-grained visual recognition and image understanding. Meanwhile, the results demonstrate that OVPG possesses the ability to continuously refresh benchmarks, ensuring adaptability as model capabilities continue to evolve.
\end{itemize}

\section{Related Work}

% Current benchmarks for LMMs mainly focus on the following five categories: perception, reasoning, specific domains, key capacities, and modalities \cite{li2024survey,huang2024survey}. 

\subsection{Evaluating Large Multimodal Models}
Designing holistic and objective benchmarks for evaluating LMMs is essential for investigating the strengths and weaknesses of LMMs, which serve as a basis for improving the development of next-generation LMMs. 
Early benchmarks such as TextVQA \cite{singh2019towards}, DocVQA \cite{mathew2021docvqa}, and GQA \cite{hudson2019gqa} are all single-task-oriented, {\it e.g.}, OCR-related understanding tasks with visual question answering (VQA). Later, more benchmarks, including MM-Vet \cite{yu2024mm}, MMBench \cite{liu2024mmbench}, SEED-Bench \cite{li2024seed}, and LVLM-EHub \cite{xu2024lvlm}, were proposed to provide more comprehensive evaluations of the overall capabilities ({\it e.g.}, perception, reasoning, specific domains, hallucination, and modalities) of LMMs \cite{li2024survey,huang2024survey}. 
Meanwhile, a number of interdisciplinary benchmarks have also been proposed to verify the wide range of applications of LMMs. For example, MMMU \cite{yue2024mmmu} collects massive multi-discipline questions that involve college-level subject knowledge and deliberate reasoning.
OBI-Bench \cite{chen2024obi} examines the coarse- and fine-grained perception ability of LMMs in different oracle bone character processing tasks. GMAI-MMBench \cite{ye2024gmai} and M3D \cite{bai2024m3d} were proposed to evaluate the capacities of LMMs in medical image analysis and investigate whether they can offer substantial assistance for diagnosis and treatment. DriveLM \cite{sima2024drivelm} presents a graph VQA form to evaluate the perception, prediction, and planning abilities of LMMs in autonomous driving.

\subsection{Data Contamination}
Data contamination is a critical issue encountered during the training and evaluation of both LLMs and LMMs, which can significantly impact the model’s performance estimation, leading to unauthentic results that do not accurately reflect its true capabilities \cite{deng2023investigating,xu2024benchmark}. Recent studies \cite{lovin2023gpt,kocon2023chatgpt,golchintime,li2024task, li2023open} have revealed the existence of data contamination problems in LLMs including GPT, LLaMA, and Vicuna series. 
Golchin and Surdeanu \cite{golchin2023data} proposed a data contamination quiz to detect data contamination in LLMs and estimate the amount of it.
Authors in \cite{yang2023rethinking} observed such contamination in widely used benchmarks such as
MMLU \cite{hendrycksmeasuring}, GSK8k \cite{cobbe2021training}, and HumanEval \cite{chen2021evaluating}, and proposed an LLM-based decontamination method by accurately removing rephrased samples.
Yang {\it et al.} \cite{yang2024dynamic} investigated the severe overlap of images with pre-training data in existing LMM benchmarks.
Consequently, it becomes imperative for the research community to develop methods for avoiding or mitigating the impact of data contamination in LMM benchmark construction.

\begin{figure}[t]
  \centering
  \includegraphics[width=0.98\linewidth]{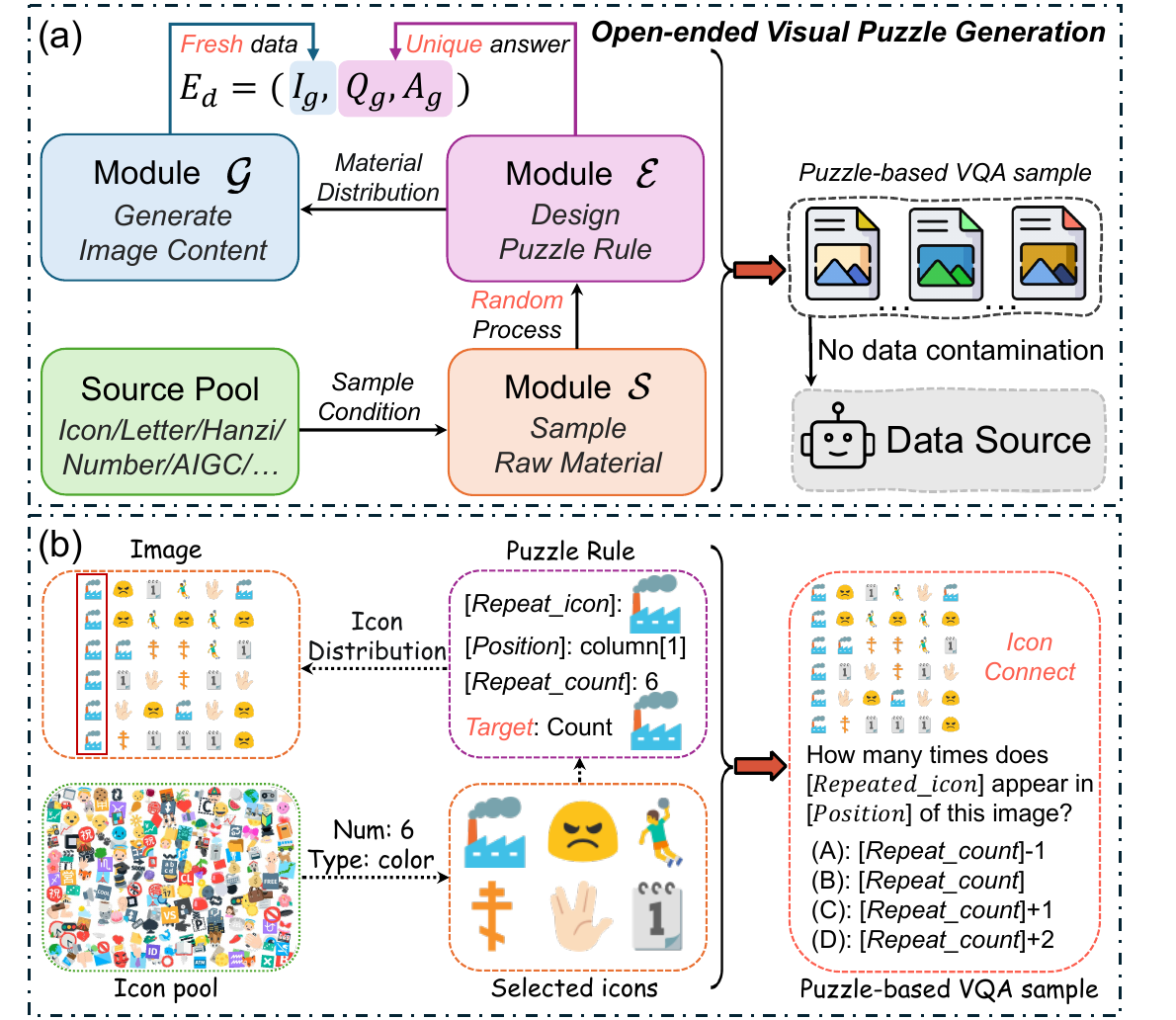}
   \caption{Illustration of the proposed fully dynamic multimodal evaluation framework and the design pipeline of a specific puzzle task. (a) The Open-ended Visual Puzzle Generation (OVPG) framework. (b) The pipeline for generating Icon Connect VQA samples.}
   \label{framework}
\end{figure}

\begin{table*}[!t]
    \centering
    \renewcommand\arraystretch{1.08}
    \caption{Overview of the six puzzle tasks in \textbf{PuzzleBench}. We report the image sources, design conditions, output targets as well as the query-answer pairs of each task. IC, HM, WS, GS, JS, and DH denote icon connect, hanzi matrix, word search, grid sum, Jigsaw, and difference hunt tasks, respectively.}
    \vspace{-6pt}
    \resizebox{.85\linewidth}{!}{
    \begin{tabular}{llcccccll}
        \toprule
        \multirow{2}{*}{\textbf{Field}} & \multirow{2}{*}{\textbf{Task}} & \multicolumn{2}{c}{\textbf{Image}} & \multirow{2}{*}{\textbf{Source}} & \multirow{2}{*}{\textbf{Condition $C$}} & \multirow{2}{*}{\textbf{Target}} & \multicolumn{2}{c}{\textbf{Description}} \\
        \cmidrule{3-4} \cmidrule{8-9}
         & & \textbf{\# Grid} & \textbf{Size} & & & & \textbf{Query} & \textbf{Answer} \\
         \midrule
      \multirow{2}{*}{\makecell[l]{\\Visual\\Recognition}}&IC&[3, 9]&$512^2$&>150 icon sets&\makecell[c]{Class \& \\ Color}&Number&\makecell[l]{How many times\\ does [icon] appear \\in [position] of \\this image?}&\makecell[l]{(A): $m-1$ \\(B): $m$\\(C): $m+1$\\(D): $m+2$}\\
      \cdashline{2-9}
       &HM&[3, 9]&$512^2$&\makecell[l]{>20k hanzi \& \\ 500 near-form}&\makecell[c]{Class \&\\ Near-form}&Number&\makecell[l]{How many unique\\Chinese characters are\\in the grid image}&\makecell[l]{(A): $m-1$ \\(B): $m$\\(C): $m+1$\\(D): $m+2$}\\
       \midrule
       \multirow{2}{*}{\makecell[l]{\\Logical\\Reasoning}}&WS&[3, 9]&$512^2$&>3k words&\makecell[c]{Class \& \\ Length}&\makecell[c]{Binary\\ judgment}&\makecell[l]{Does this grid image\\ contain the [word]?}&\makecell[l]{(A): Yes \\(B): No}\\
       \cdashline{2-9}
       &GS&[3, 9]&$512^2$&Integers &\makecell[c]{Operation \& \\ Value}&Number&\makecell[l]{What is the sum \\of all
the numbers\\ in the grid image?}&\makecell[l]{(A): $[sum]-10$ \\(B): $[sum]$\\(C): $[sum]+10$\\(D): $[sum]+20$}\\
      \midrule
       \multirow{2}{*}{\makecell[l]{\\Context\\Understanding}}&JS&[3, 9]&$256^2$-$1024^2$&AIGC&Prompt&Tile id&\makecell[l]{Which tile is\\ missing 
from the \\incomplete image?}&\makecell[l]{(A): target tile \\(B): tile 0\\(C): tile 1\\(D): tile 2}\\
       \cdashline{2-9}
       &DH&\textendash&$256^2$-$1024^2$&AIGC&Prompt&Number&\makecell[l]{How many differences \\ are there between\\ these two images?}&\makecell[l]{(A): $m-1$ \\(B): $m$\\(C): $m+1$\\(D): $m+2$}\\
    \bottomrule
\end{tabular}}
    \label{design}
\end{table*}

\subsection{Dynamic Evaluation}
To tackle the data contamination problem in the current benchmarks, some researchers explored a novel dynamic evaluation paradigm. 
Fan {\it et al.} \cite{fan2024nphardeval} built a benchmark, NPHardEval with a dynamic update mechanism, for measuring the reasoning ability of LMMs, where the datapoints are
refreshed monthly.
DyVal \cite{zhu2023dyval} proposed a graph-informed dynamic data generation method and controllable complexity to address the challenges of data contamination.
Similarly, Yang {\it et al.} \cite{yang2024dynamic} designed a vision-language bootstrapping  strategy that can dynamically generate new visual question-answering samples while ensuring that newly generated samples remain consistent with the original ones. However, such bootstrapping merely transforms existing data, without updating the data source, which we refer to as a {\it semi-dynamic evaluation}.

\section{The PuzzleBench}
This section introduces our proposed fully dynamic multimodal evaluation framework, \textbf{Open-ended Visual Puzzle Generation (OVPG)}, along with the benchmark we build upon it, \textbf{PuzzleBench}. Sec. \ref{3.1} provides a overview. Sec. \ref{3.2} details the architecture and components of OVPG. Sec. \ref{sec 3.3} describes how PuzzleBench evaluates different dimensions of LMM capabilities, introduces the six puzzle tasks, and explains the full construction process of the benchmark. PuzzleBench serves as a concrete instantiation of OVPG, demonstrating its ability to generate fully dynamic, fine-grained, and capability-aligned VQA samples.

\subsection{Overview}\label{3.1}

\subsubsection{Definition of Fully Dynamic Multimodal Evaluation} A fully dynamic multimodal evaluation for LMMs implies that the evaluation samples consisting of images, questions, and corresponding ground-truth answers are dynamically generated and vary across instances. Formally, let a static multimodal evaluation sample be defined as: $E_s=(I, Q, A)$ where $I$, $Q$ and $A$ denote the image, question, and ground truth answer. In contrast, a fully dynamic evaluation sample is defined as $E_d=(I_g, Q_g, A_g)$, where the subscript $g$ indicates dynamic generation. Dynamic samples $E_d$ have no overlap with static samples $E_s$, ensuring unbiased and reliable evaluation.
 
\subsubsection{Our Approach} To address the critical challenges posed by static benchmarks and data contamination issues, as shown in Fig. \ref{framework}(a), we propose \textbf{Open-ended Visual Puzzle Generation (OVPG)}, a fully dynamic multimodal evaluation framework tailored for assessing LMMs. OVPG leverages a three-stage pipeline composed of a \textit{raw material sampling module}, $\mathcal{S}$, \textit{puzzle rule design module} $\mathcal{E}$ , and a \textit{visual content generation module} $\mathcal{G}$ to construct evaluation samples on the fly. The generation process is highly \textbf{randomized}, introducing stochasticity through dynamic sampling and different open-ended puzzle rules, ensuring that no two instances are alike. As a result, the generated images are inherently \textbf{fresh}, drawn from diverse source pools and synthesized automatically. Moreover, every puzzle instance forms a \textbf{unique} query-answer pair, where the ground-truth answer is exclusively and deterministically derived from the generation process, ensuring unambiguous correctness in evaluation. This design enables OVPG to produce scalable, contamination-free, and dynamically evolving evaluation benchmarks that align with the rapid advancement of LMMs.

\subsection{Open-ended Visual Puzzle Generation}\label{3.2}
\subsubsection{Raw Material Sampling Module}
The raw material sampling module $\mathcal{S}$ introduces stochasticity into the evaluation pipeline by randomly selecting raw material from diverse and extensive source pools, based on predefined sampling conditions ${C}$. Formally, this sampling process is expressed as:
\vspace{-1pt}
\begin{equation}
V = \mathcal{S}(\mathcal{R}, C) 
\end{equation}
Here, $\mathcal{R}$ represents the comprehensive source pool, and $V$ denotes the sampled raw material. These source pools consist of semantically meaningful visual categories, such as alphabets or words from various languages, icon sets, and rapidly expanding AI-generated images. The sampling conditions ${C}$ specify what kinds of raw material should be selected ({\it e.g.}, category, appearance, quantity), serving as constraints for each puzzle instance. This module provides \textbf{randomness} in image content selection under explicit conditions.

\subsubsection{Puzzle Rule design module}
The puzzle rule design module $\mathcal{E}$ is an open-ended component responsible for defining the core objective of a puzzle instance. It first determines the specific puzzle target $T$ ({\it e.g.}, identify repetitions, detect differences, compute sums), which guides the downstream construction process. Based on the established target, $\mathcal{E}$ then defines the arrangement and distribution patterns of the sampled raw material, including its spatial layout, frequency of appearance, repetition constraints, and other task-specific rules. Formally, this puzzle rule design is expressed as:
\begin{equation}
R = \mathcal{E}(V, T) 
\end{equation}
This module enables diverse and flexible puzzle rule instantiations tailored to each puzzle type.
Each puzzle instance forms a question-answer pair, derived from the sampled raw material and puzzle rules. This module ultimately ensures the \textbf{uniqueness} and verifiability of the ground-truth answer.

\subsubsection{Visual Content Generation Module}

The visual content generation module $\mathcal{G}$ takes the sampled raw material and generates visual content according to the distribution parameters specified in the puzzle rules $R$. Formally, this generation process is defined as:
\begin{equation}
I_G = \mathcal{G}(V, R)
\end{equation}
where $I_G$ denotes the generated puzzle image. This module ensures that the generated visual content are visually diverse and structurally valid, while maintaining a high degree of \textbf{freshness}.

\subsection{Constuction of PuzzleBench}
\label{sec 3.3}
\subsubsection{Dimensions and Puzzle Tasks}

To comprehensively evaluate LMMs across critical cognitive dimensions, PuzzleBench carefully selects puzzle tasks based on their ability to assess visual recognition, logical reasoning, and context understanding. Each puzzle type is chosen for its clear alignment with these dimensions and its suitability for dynamic generation from the open-ended puzzle design.
We identify three core competency dimensions for LMMs:

\noindent\textbf{Visual Recognition}: the ability to accurately perceive, localize, and identify key visual elements within an image.

\noindent\textbf{Logical Reasoning}: the ability to integrate visual recognition with logical inference, enabling the identification of relational structures and the application of reasoning over spatial or symbolic patterns.

\noindent\textbf{Context Understanding}: the ability to comprehend semantic information embedded in the image, with a focus on spatial configurations and contextual associations among visual components.

To evaluate the three competency dimensions described above, we design a set of six puzzle tasks within our framework. Specifically, the visual recognition dimension includes:

\noindent\textbf{Icon Connect}: identifying recurring icons in an image grid.

\noindent\textbf{Hanzi Matrix}: counting distinct hanzi in an image grid.\\
The logical reasoning dimension includes:

\noindent\textbf{Word Search}: locating target words in a letter-based image grid.

\noindent\textbf{Grid Sum}: computing the total of digits in a numeric image grid.\\
The context understanding dimension includes:

\noindent\textbf{Jigsaw}: selecting the missing tile to complete an image grid.

\noindent\textbf{Difference Hunt}: spotting differences between paired images.

\begin{table*}[t]  % \small
    \centering
    \renewcommand\arraystretch{1}
    \caption{Full evaluation results of 14 LMMs on \textbf{PuzzleBench} under the \textit{Fixed Option} experiment. The best performance is marked in \textbf{bold} and the second-best performance is \underline{underlined}, reported separately for both open-source and proprietary LMMs. \textcolor{red}{Red} indicates LMMs whose performance falls below the random guess baseline.}
    \vspace{-5pt}
    \resizebox{.9\linewidth}{!}{\begin{tabular}{l|cc|cc|cc|c}
    \toprule
    {\textbf{Field}}  & \multicolumn{2}{c|}{{Visual Recognition}} & \multicolumn{2}{c|}{{Logical Reasoning}} & \multicolumn{2}{c|}{{Context Understanding}} & \multirow{2}{27pt}{\textit{Average}}\\ \cdashline{1-7}
     \textbf{LMM / Puzzle}  &{\textit{Icon Connect}} & {\textit{Hanzi Matrix}} & {\textit{Word Search}} & \textit{Grid Sum} & {\textit{Jigsaw}} & {\textit{Difference Hunt}} & \\ \hline 
     \multicolumn{8}{l}{\textbf{\textit{Open-source}}} \\ \cdashline{1-8}
    Qwen2.5-VL-3B & \textcolor{red}{0.208} & \textcolor{red}{0.212} & \textcolor{red}{0.339} & \textcolor{red}{0.002} & 0.268 & \textcolor{red}{0.213} & \textcolor{red}{0.207}\\
    Qwen2.5-VL-7B & \textcolor{red}{0.214} & \textcolor{red}{0.205} & \textcolor{red}{0.366} & \textcolor{red}{0.050} & 0.288 & 0.236 & \textcolor{red}{0.226}\\ 
    Qwen2.5-VL-72B & 0.365 & 0.336  & \textbf{0.576} & \textbf{0.622} & \textbf{0.433} & 0.260 & \underline{0.432}\\ 
    Qwen2-VL-2B & \textcolor{red}{0.171} & \textcolor{red}{0}  & \textcolor{red}{0.195} & \textcolor{red}{0.098} & \textcolor{red}{0.202} & \textcolor{red}{0} & \textcolor{red}{0.111}\\
    Qwen2-VL-7B & \textcolor{red}{0.214} & \textcolor{red}{0.173} &\textcolor{red}{0.267} & \textcolor{red}{0.115} & \underline{0.368} & \textcolor{red}{0} &  \textcolor{red}{0.189}\\ 
    InternVL2.5-8B & 0.361 & 0.287 & \textcolor{red}{0.381} & 0.342 & \textcolor{red}{0.237} & \underline{0.383} & 0.332\\
    InternVL2.5-38B & \underline{0.428} & \underline{0.422} & \textcolor{red}{0.475} & 0.316 & 0.259 & 0.314 & 0.369\\
    InternVL2.5-78B & \textbf{0.506} & \textbf{0.502} & \underline{0.556} & \underline{0.529} & 0.256 & \textbf{0.465} & \textbf{0.469}\\ 
    DeepSeek-VL2-small & \textcolor{red}{0.245} & \textcolor{red}{0.223} & \textcolor{red}{0.099} & \textcolor{red}{0.215} & \textcolor{red}{0.230} & 0.298 & \textcolor{red}{0.218}\\ 
    Llava-OneVision-7B & \textcolor{red}{0} & \textcolor{red}{0} & \textcolor{red}{0} & \textcolor{red}{0} & \textcolor{red}{0} & \textcolor{red}{0} & \textcolor{red}{0}\\ \hline
    \multicolumn{8}{l}{\textbf{\textit{Proprietary}}} \\ \cdashline{1-8}
    Qwen-VL-Max-1220 & \textbf{0.534} & 0.363 & \textcolor{red}{0.323} & \textbf{0.706} & 0.370 & 0.297 & \underline{0.432}\\
    Gemini-1.5-Pro & 0.339 & \underline{0.491} & \underline{0.568} & 0.288 & \textbf{0.486} & \underline{0.403} & 0.429\\
    Gemini-2.0-Flash  & \underline{0.365} & \textbf{0.541} & \textbf{0.652} & 0.549 & \underline{0.481} & \textbf{0.524} & \textbf{0.519}\\
    GPT-4o & 0.363 & 0.362 & 0.540 & \underline{0.668} & 0.464 & \textcolor{red}{0.193} &  \underline{0.432}\\ \hline
    \textit{Random guess} & 0.253 & 0.254 & 0.495 & 0.236 & 0.256 & 0.231 & 0.288\\ 
     \bottomrule
    \end{tabular}}
    %\vspace{-13pt}
    \label{exp1}
\end{table*}

\subsubsection{Puzzle Design}

PuzzleBench consists of 11,840 automatically generated puzzle-based VQA samples, specifically designed to cover the three core competency dimensions described above. Each dimension features two puzzle tasks, yielding six distinct puzzle tasks with different \textit{difficulty levels}, achieved by controlling parameters such as grid size, image resolution, or specific puzzle rules. Tab. \ref{design} provides detailed design specifications for each puzzle task, including sampling condition, puzzle rule, and question-answer format.

\noindent\textbf{Icon Connect and Hanzi Matrix.} Both tasks produce 2,000 grid images each (4,000 total), primarily testing \emph{visual recognition}. In Icon Connect, icons are sampled from the large-scale \textit{Iconify} dataset, ensuring a variety of styles and densities. Hanzi Matrix samples Chinese characters (hanzi) from a source pool of over 20,000 characters and 500 visually similar sets. For both tasks, difficulty scales with the grid size $N$ and specific puzzle rules ({\it e.g.}, enforcing consecutive icons in Icon Connect, introducing visually similar hanzi in Hanzi Matrix). Fig. \ref{framework}(b) show the pipeline of Icon Connect.

\noindent\textbf{Word Search and Grid Sum.} Each of these tasks also generates 2,000 grid images (4,000 total), focusing on \emph{logical reasoning}. Word Search presents an $N \times N$ letter grid where a target word, sampled from a corpus of 3,000+ English words, must be identified. All other cells are randomly populated with letters, making text detection and spatial alignment more challenging at larger $N$. Grid Sum requires models to compute the sum of values randomly sampled from $[-100, 100]$, including decimals. Difficulty in both tasks depends on grid size $N$ and specific puzzle rules ({\it e.g.}, target word placement or numeric distribution).

\noindent\textbf{Jigsaw and Difference Hunt.} Together, these tasks yield 3,840 samples (1,920 each) and emphasize \emph{context understanding}. Both leverage AI-generated images from 12 topic categories, produced via Stable Diffusion 3.5 using GPT-4 generating prompts. In Jigsaw, difficulty is driven by image resolution ($256^2$, $512^2$, $1024^2$) and the number of tiles ({\it i.e.}, splitting the image into $N \times N$ segments), requiring identification of a missing tile. Difference Hunt presents two nearly identical images with local distortions ({\it e.g.}, blur, noise, overexposure). Its difficulty further depends on image size and a difficulty level $l$ (ranging 1–5), challenging models’ capacity to detect increasingly subtle differences.

All puzzle tasks adopt a multiple-choice format. Most puzzles provide four answer options, except for certain tasks ({\it e.g.}, Word Search) that require a binary Yes/No response. Overall, the puzzle VQA generation in OVPG endows PuzzleBench with randomized process, fresh image source, and uniquely determined ground-truth answers, thoroughly assessing models’ recognition, reasoning, and context understanding. Additional design details and difficulty level  can be found in the supplementary materials.

\section{Experiments}
\subsection{Experimental Settings}

{\bf Benchmark Candidates.} 14 up-to-date and competitive LMMs are selected for evaluation, which consists of 4 proprietary LMMs and 10 open-source LMMs:
\begin{itemize}
    \item The proprietary LMMs include GPT-4o \cite{GPT-4o}, Gemini-2.0 flash \cite{gemini2}, Gemini-1.5-Pro \cite{reid2024gemini}, and Qwen-VL-Max \cite{bai2023qwen}.
    \item The open-source LMMs include Qwen2.5-VL series (3B, 7B, and 72B) \cite{bai2025qwen25vl}, Qwen2-VL series (2B and 7B) \cite{wang2024qwen2}, InternVL2.5 series (8B, 38B, and 78B) \cite{chen2024expanding}, DeepSeek-VL2-small \cite{wu2024deepseek}, and Llava-OneVision-7B \cite{li2024llava}.
\end{itemize}

\noindent
{\bf Implementation Details.} To systematically examine and mitigate potential biases introduced by answer-option arrangements in our experiments, we designed and conducted evaluations under three distinct experimental conditions: \\
% \begin{itemize}
{Fixed Option: }For the VQA samples, the correct answers were consistently positioned as option <B> across all trials. Similarly, in Word Search, the correct answer was uniformly set to option <A> ("Yes"). This controlled setting allowed us to observe the effect of predictable answer positions on model performance.\\
{Randomized Option: } Unlike Fixed Option, we randomized the placement of correct answers within the provided options. For VQA tasks, ground-truth answers were randomly assigned among options <A>, <B>, <C>, and <D>. In Word Search, "Yes" and "No" responses were randomly shuffled between options <A> and <B>. This randomization aimed to neutralize positional biases and better represent realistic evaluation conditions.\\
{Direct Answer: } In this experimental setup, LMMs were prompted to generate answers directly without provided multiple-choice options. However, since the Jigsaw task inherently requires selecting among four image-based options, it was not included in the direct answer experiment. 
% \end{itemize}
Moreover, all evaluations are conducted in a zero-shot setting for a fair comparison.

\noindent
{\bf Evaluation Criteria.} We calculate the \textit{accuracy} of verbal answers to multiple-choice questions using exact matching as our primary evaluation metric. In the direct answer experiment, we additionally compute the \textit{Mean Relative Error (MRE)} to quantify the deviation between the LMM answers and the unqiue ground-truth answers, This metric is particularly suitable for our PuzzleBench, where most tasks have ground-truth answers in the form of numerical values, thereby providing a more fine-grained assessment of answer quality beyond categorical correctness.
\begin{equation}
\text{MRE} = \frac{1}{N} \sum_{i=1}^{N} \frac{|\hat{y}_i - y_i|}{|y_i|}    
\end{equation}
where \( N \) denotes the number of evaluated samples in each puzzle task, \( \hat{y}_i \) represents the predicted value of LMMs for the \( i \)-th sample, and \( y_i \) is the corresponding ground-truth answer.
\subsection{Main Results}

In Tab. \ref{exp1}, Tab. \ref{exp2} and Tab. \ref{exp3}, we present the experimental results corresponding to the Fixed Option, Random Option, and Direct Answer settings, respectively. We evaluate three core competencies of LMMs, namely visual recognition, logical reasoning, and context understanding, using six puzzle tasks from the PuzzleBench. We summarize key findings as below.

\begin{table*}[t]  %\small
    \centering
    \renewcommand\arraystretch{1}
    \caption{Evaluation results of 10 LMMs on \textbf{PuzzleBench} under the \textit{Randomized Option} experiment. The best performance is marked in \textbf{bold} and the second-best performance is \underline{underlined}, reported separately for both open-source and proprietary LMMs. \textcolor{red}{Red} indicates LMMs whose performance falls below the random guess baseline.}
    \vspace{-5pt}
    \resizebox{.9\linewidth}{!}{\begin{tabular}{l|cc|cc|cc|c}
    \toprule
    {\textbf{Field}}  & \multicolumn{2}{c|}{{Visual Recognition}} & \multicolumn{2}{c|}{{Logical Reasoning}} & \multicolumn{2}{c|}{{Context Understanding}} & \multirow{2}{27pt}{\textit{Average}}\\ \cdashline{1-7}
     \textbf{LMM / Puzzle}  &{\textit{Icon Connect}} & {\textit{Hanzi Matrix}} & {\textit{Word Search}} & \textit{Grid Sum} & {\textit{Jigsaw}} & {\textit{Difference Hunt}} & \\ \hline 
     \multicolumn{8}{l}{\textbf{\textit{Open Source}}} \\ \cdashline{1-8}
    Qwen2.5-VL-3B & \textcolor{red}{0.252} & 0.255 & \textcolor{red}{0.245} & \textcolor{red}{0.020} & \textbf{0.268} & 0.285 & \textcolor{red}{0.221}\\
    Qwen2.5-VL-7B & 0.265 & \textcolor{red}{0.228} & \textcolor{red}{0.368} & \textcolor{red}{0.109} & \textcolor{red}{0.226} & 0.304 & \textcolor{red}{0.250}\\ 
    Qwen2.5-VL-72B & \underline{0.412} & \underline{0.400}  & \underline{0.593} & \underline{0.370} & \textcolor{red}{0.233} & 0.257 & \underline{0.377}\\ 
    InternVL2.5-8B & 0.375 & 0.357 & \textcolor{red}{0.368} & 0.236 & \textcolor{red}{0.248} & \underline{0.329} & 0.319\\
    InternVL2.5-78B & \textbf{0.457} & \textbf{0.408} & \textbf{0.646} & \textbf{0.377} & \textcolor{red}{0.171} & \underline{0.434} & \textbf{0.415}\\ 
    DeepSeek-VL2-small & 0.279 & 0.258 & \textcolor{red}{0.279} & \textcolor{red}{0.234} & \textcolor{red}{0.217} & \textcolor{red}{0.196} & \textcolor{red}{0.244}\\ \hline
    \multicolumn{8}{l}{\textbf{\textit{Proprietary}}} \\ \cdashline{1-8}
    Qwen-VL-Max-1220 & \textbf{0.469} & 0.294 & \textcolor{red}{0.335} & \underline{0.369} & \textbf{0.336} & \textcolor{red}{0.184} & 0.331\\
    Gemini-1.5-Pro & 0.403 & \underline{0.420} & \underline{0.526} & \textbf{0.379} & \underline{0.329} & \underline{0.366} & \underline{0.404}\\
    Gemini-2.0-Flash  & \underline{0.420} & \textbf{0.673} & \textbf{0.706} & 0.353 & 0.255 & \textbf{0.493} & \textbf{0.483}\\
    GPT-4o & 0.314 & 0.282 & 0.514 & 0.258 & 0.315 & \textcolor{red}{0.179} & 0.310\\ \hline
    \textit{Random guess} & 0.254 & 0.254 & 0.511 & 0.253 & 0.254 & 0.248 & 0.296\\ 
     \bottomrule
    \end{tabular}}
    \vspace{-4pt}
    \label{exp2}
\end{table*}

\noindent{\bf Gap between Open-source and Proprietary Models.} 
There exists a notable performance gap between open-source and proprietary LMMs across all six puzzle tasks and three evaluation dimensions. Most open-source models perform only marginally better than random guessing. Interestingly, however, InternVL2.5 78B manages to keep pace with proprietary models, achieving a competitive overall accuracy of 46.9$\%$. Moreover, several smaller-scale open-source models, such as Qwen2-VL(3B and 7B), and Llava-OneVision-7B, even underperform compared to the random guess baseline. In contrast, all proprietary models consistently outperform the random baseline, with Gemini-2.0-Flash achieving the best overall performance, reaching an average accuracy of 51.9$\%$. \\
\noindent{\bf Model Performance in Different Dimensions.}
Across the three dimensions of PuzzleBench, we observe distinct differences in capabilities among proprietary LMMs, large-scale open-source LMMs, and smaller-scale open-source LMMs in terms of \textit{visual recognition, logical reasoning, and context understanding.} Proprietary LMMs demonstrate the strongest performance in \textit{logical reasoning}, suggesting that they are particularly proficient at tasks involving simple OCR and goal-oriented reasoning. A similar trend can be observed in large open-source LMMs, such as Qwen2.5-VL 72B and InternVL2.5 78B, which also perform relatively well in this dimension. In contrast, smaller-scale open-source models show comparable performance across all three dimensions, with results only marginally above, or sometimes even below, the \textit{random guess} baseline. This indicates that these models lack sufficient capabilities in all aspects of dimensions evaluated in PuzzleBench.\\
\noindent{\bf Model Performance in Different Tasks.}
The puzzles in PuzzleBench provide a multimodal evaluation of current LMMs, spanning a range of visual granularities and task types. From the performance of proprietary LMMs, we observe that existing models still struggle with complex visual OCR and fine-grained semantic understanding of images. For instance, GPT-4o significantly outperforms others on Word Search and Grid Sum, both of which involve visually simple elements such as English characters and digits. In contrast, tasks like Icon Connect and Hanzi Matrix feature puzzle images composed of more complex visual elements, such as intricate icons and Chinese characters (hanzi), which pose greater challenges for visual recognition and understanding. A similar result is observed in large-scale open-source LMMs. The substantial performance disparity across tasks highlights a key limitation of current models, their insufficient ability to process and reason over fine-grained visual semantics. These findings underscore the pressing need for further advancements in fine-grained visual perception and context-aware image understanding in LMMs.

In conclusion, PuzzleBench provides a comprehensive evaluation of LMMs across three core dimensions: visual recognition, logical reasoning, and context understanding. Results in Table 1 show that Gemini-2.0-Flash achieves the highest overall performance. However, the general performance across all models reveals substantial room for improvement, particularly in tasks requiring fine-grained visual perception and image understanding.

\begin{table*} \small
    \centering
    \renewcommand\arraystretch{1}
    \caption{Evaluation results of 10 LMMs on \textbf{PuzzleBench} under the \textit{Direct Answer} experiment. The best performance is marked in \textbf{bold} and the second-best performance is \underline{underlined}, reported separately for both open-source and proprietary LMMs. The metrics used in this experiment are Accuracy and Mean Relative Error (MRE).}
    \vspace{-5pt}
    \resizebox{.9\linewidth}{!}{\begin{tabular}{l|cc|cc|cc|c}
    \toprule
    {\textbf{Field}}  & \multicolumn{2}{c|}{{Visual Recognition}} & \multicolumn{2}{c|}{{Logical Reasoning}} & \multicolumn{2}{c|}{{Context Understanding}} & \multirow{2}{27pt}{\textit{Average}}\\ \cdashline{1-7}
     \textbf{LMM / Puzzle}  &{\textit{Icon Connect}} & {\textit{Hanzi Matrix}} & {\textit{Word Search}} & \textit{Grid Sum} & {\textit{Jigsaw}} & {\textit{Difference Hunt}} & \\ \hline 
     \multicolumn{8}{l}{\textbf{\textit{Open Source}}} \\ \cdashline{1-8}
    Qwen2.5-VL-3B & 0.1510/0.7584 & 0.0110/1.4767 & 0.1720 & 0/1.7126 & NA & 0.0379/0.5924 & 0.0745/1.1350\\
    Qwen2.5-VL-7B & 0.1305/0.6715 & 0.1720/0.2598 & 0.2390 & 0/1.3222 & NA & 0.0672/0.4237 & 0.1217/0.6693\\ 
    Qwen2.5-VL-72B & \textbf{0.3490}/\textbf{0.2929} & \underline{0.3453}/\textbf{0.2098}  & \underline{0.4930} & 0/1.1608 & NA & \underline{0.2340/0.3218} & \underline{0.2843/0.4963}\\ 
    InternVL2.5-8B & 0.2140/0.4262 & 0.1915/0.7792 & 0.291 & 0/\underline{1.1420} & NA & 0.1534/0.4260 & 0.1670/0.6934\\
    InternVL2.5-78B & \underline{0.3370}/\underline{0.3129} & 0.3200/\underline{0.2136} & \textbf{0.515} & \textbf{0.0020/0.6035} & NA & \textbf{0.3191/0.2575} & \textbf{0.2986/0.3469}\\ 
    DeepSeek-VL2-small & 0.2720/0.4745 & \textbf{0.3620}/0.3784 & 0.1155 & \underline{0.0005}/2.8198 & NA & 0.1231/0.5871 & 0.1746/1.0650\\ \hline
    \multicolumn{8}{l}{\textbf{\textit{Proprietary}}} \\ \cdashline{1-8}
    Qwen-VL-Max-1220 & 0.3190/0.3731 & 0.4490/0.1686 & 0.382 & 0/1.2142 & NA & 0.3134/0.2838 & 0.2927/0.5099\\
    Gemini-1.5-Pro & \textbf{0.4115}/\textbf{0.2581} & \underline{0.4930/0.1259} & 0.565 & 0.0010/\underline{0.9327} & NA & 0.4044/\textbf{0.1749} & \underline{0.3750}/\textbf{0.3729}\\
    Gemini-2.0-Flash  & \underline{0.3660}/\underline{0.3618} & \textbf{0.6250/0.0908} & \textbf{0.654} & \underline{0.0375}/\textbf{0.8891} & NA & \textbf{0.4773}/0.2033 & \textbf{0.4320}/\underline{0.3863}\\
    GPT-4o & 0.1480/0.4657 & 0.3055/0.2150 & \underline{0.598} & \textbf{0.0725}/1.3338 & NA & \underline{0.4631/0.1799} & 0.31742/0.5486\\ 
     \bottomrule
    \end{tabular}}
    %\vspace{-13pt}
    \label{exp3}
\end{table*}

\subsection{Impact of Query-answer Format}

To further investigate the effect of answer-option positions, we conducted a \textit{randomized option} experiment with 10 LMMs, and the results are presented in Tab. \ref{exp2}. We observe that most models experience a noticeable performance drop when transitioning from fixed to random options. In the \textit{fixed option} experiment, Gemini-2.0-Flash achieves the best overall performance. However, under the randomized option condition, its average score decreases by 3.6\%. Similarly, InternVL2.5-78B, the top-performing open-source model, experiences a 5.4\% decline. These findings underscore the influence of answer-position biases on model performance and highlight the vulnerability of current LMMs to shifts in option arrangements. This indicates that a portion of the models’ performance under fixed-option settings may stem from learned positional cues or biases, rather than solely from genuine comprehension or reasoning.

\begin{figure}[t]
  \centering
  \includegraphics[width=0.92\linewidth]{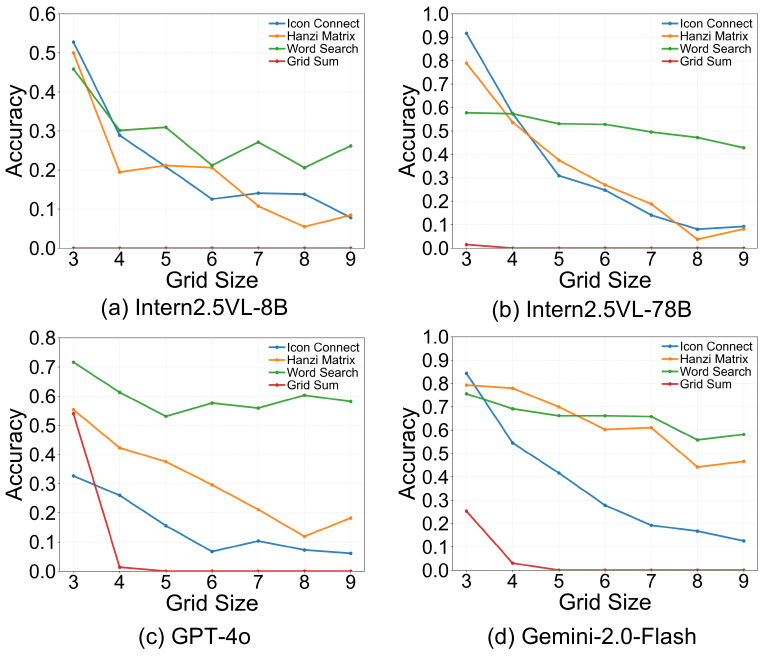}
    \caption{Accuracy of 4 LMMs across grid sizes [3, 9] on four grid-based puzzle tasks. All models exhibit a general decline in performance as the grid size increases. (a) InternVL2.5-8B, (b) InternVL2.5-78B, (c) GPT-4o, (d) Gemini-2.0-Flash.}
   \label{line}
\end{figure}

We additionally design a direct answer experiment, where models provide free-form solutions rather than choosing from given options. This setup further verifies whether models have genuinely mastered puzzle-solving abilities, rather than relying on forced-choice cues. We use accuracy and mean relative error (MRE) as evaluation metrics, and Tab. \ref{exp3} presents the corresponding results. Unsurprisingly, most LMMs exhibit even lower performance compared to the Random Option condition. Notably, all models—including proprietary ones—perform poorly on the Grid Sum puzzle when providing direct answers, with none achieving more than 10\% accuracy. The top performer, GPT-4o, only attains a 7.25\% accuracy and incurs a mean relative error of 133.38\%. 

We also plot the accuracy of 4 LMMs across varying grid sizes on four grid-based puzzle tasks, as shown in Fig. \ref{line}. As the grid size increases, all models exhibit a general decline in accuracy. In the case of InternVL2.5-8B, the accuracy curves fluctuate noticeably across grid sizes, indicating performance instability commonly observed in smaller-scale open-source models. In contrast, InternVL2.5-78B and the two proprietary LMMs, GPT-4o and Gemini-2.0-Flash, demonstrate more stable performance with smoother degradation trends. However, even in these models, there remains a substantial accuracy gap between grid size 3-9. This substantial performance gap reveals that current models still lack robust capabilities for complex recognition and reasoning without external prompts or guidance.

\begin{figure}[t]
  \centering
  \includegraphics[width=0.94\linewidth]{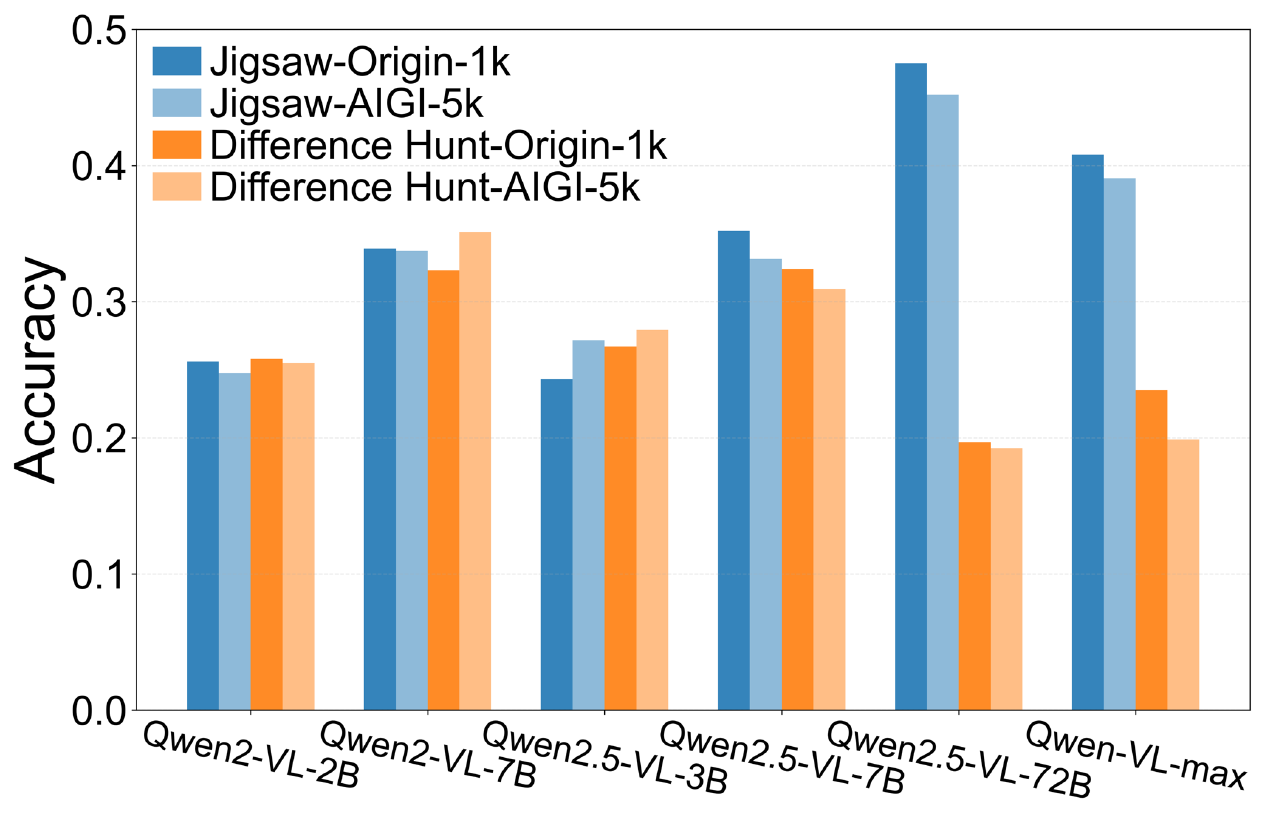}
    \caption{Accuracy of the Qwen series across four datasets: Jigsaw-Origin-1k, Jigsaw-AIGI-5k, Difference Hunt-Origin-1k, and Difference Hunt-AIGI-5k.  }
   \label{refresh}
\end{figure}

\subsection{Existing Benchmark Refreshing}
To fully leverage a large-scale static benchmark and validate both the refreshing capability of our OVPG framework and its effectiveness in dynamically generating VQA samples, we sample 1,000 images from the MSCOCO test set \cite{lin2014microsoft}, which is commonly used in the pre-training of large-scale models. Each image is paired with its corresponding caption. Our experimental procedure consists of the following three stages:
(1) \textit{Original Image Refreshing.} Using the captions of the 1,000 sampled images (a total of 5,002 caption variants), we generate 5,002 new images with Stable Diffusion 3.5 at a resolution of $512 \times 512$.
(2) \textit{Puzzle Sample Generation.} We apply OVPG to generate puzzle-based VQA samples for the Jigsaw and Difference Hunt tasks. This is done both on the original MSCOCO images and on the newly generated AI images (AIGIs). As a result, we construct four datasets: \textit{Jigsaw-Origin-1k}, \textit{Jigsaw-AIGI-5k}, \textit{Difference Hunt-Origin-1k}, and \textit{Difference Hunt-AIGI-5k}. All samples are generated using a \textit{randomized option} setting.
(3) \textit{Model Evaluation.} We evaluate multiple models from the Qwen family, including Qwen2-VL (2B and 7B), Qwen2.5-VL (3B, 7B, and 72B), and Qwen-VL-Max-1220, across all four datasets.
As shown in Fig. \ref{refresh}, our results reveal two key findings. First, in most cases, VQA samples generated from AIGIs result in lower performance compared to those generated from original MSCOCO images. Second, for the Difference Hunt task, larger Qwen models such as Qwen2.5-VL-72B and Qwen-VL-Max underperform smaller models, which is consistent with our observations in the randomized option experiments.

In summary, these results demonstrate that OVPG can effectively refresh existing static benchmarks and provide dynamic, diverse, and up-to-date evaluation samples for LMMs.

\section{Conclusion}
In this paper, we introduced Open-ended Visual Puzzle Generation (OVPG), a fully dynamic multimodal evaluation framework capable of automatically generating fresh, and uniquely solvable VQA samples. Leveraging OVPG, we designed six distinct puzzle tasks specifically targeting three core competencies of LMMs: visual recognition, logical reasoning, and context understanding, resulting in the creation of PuzzleBench, which comprises 11,840 dynamically generated VQA samples. Through comprehensive experiments involving 14 advanced LMMs, we highlighted significant limitations of current models, particularly in fine-grained image recognition, semantic understanding, and advanced reasoning tasks. Furthermore, we demonstrated OVPG's effectiveness in refreshing existing static benchmarks, underscoring the broad applicability and versatility of our framework. We anticipate that our contributions will inspire new insights and facilitate further advancements in the evaluation and development of large multimodal models.

%\section*{Acknowledgments}

\section*{Appendices}

% \begin{acks}
% \end{acks}


\bibliographystyle{ACM-Reference-Format}
\bibliography{sample-base}

\clearpage
% \appendix
\section*{Supplementary Material}
\begin{figure*}[t]
  \centering
  \includegraphics[width=1\linewidth]{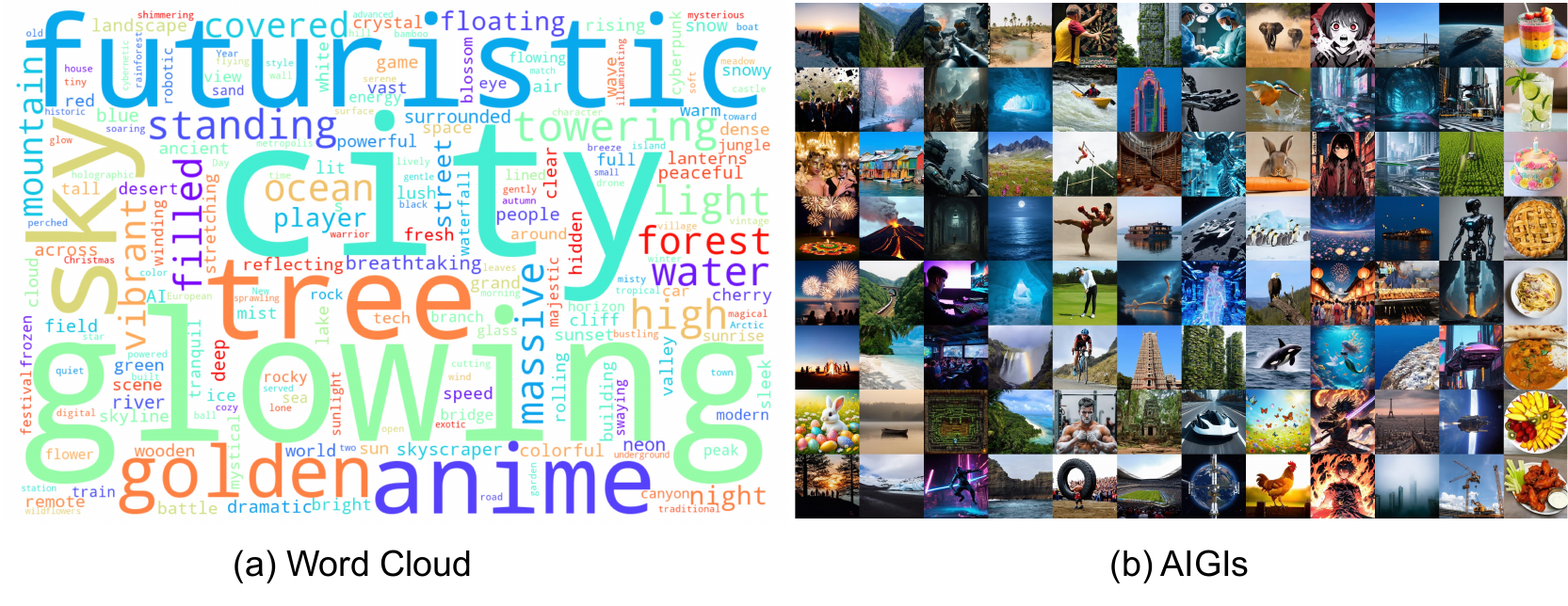}
  \caption{Overview of the prompt and visual diversity in the Jigsaw and Difference Hunt task. (a) Word cloud of GPT-4-generated prompts spanning 12 semantic categories. (b) Visual examples of AI-generated images from each category, with 8 samples randomly selected per class.}
  \label{cloud}
\end{figure*}

\section{The Design Detail of Puzzle Tasks}

\begin{figure}
  \centering
  \includegraphics[width=1\linewidth]{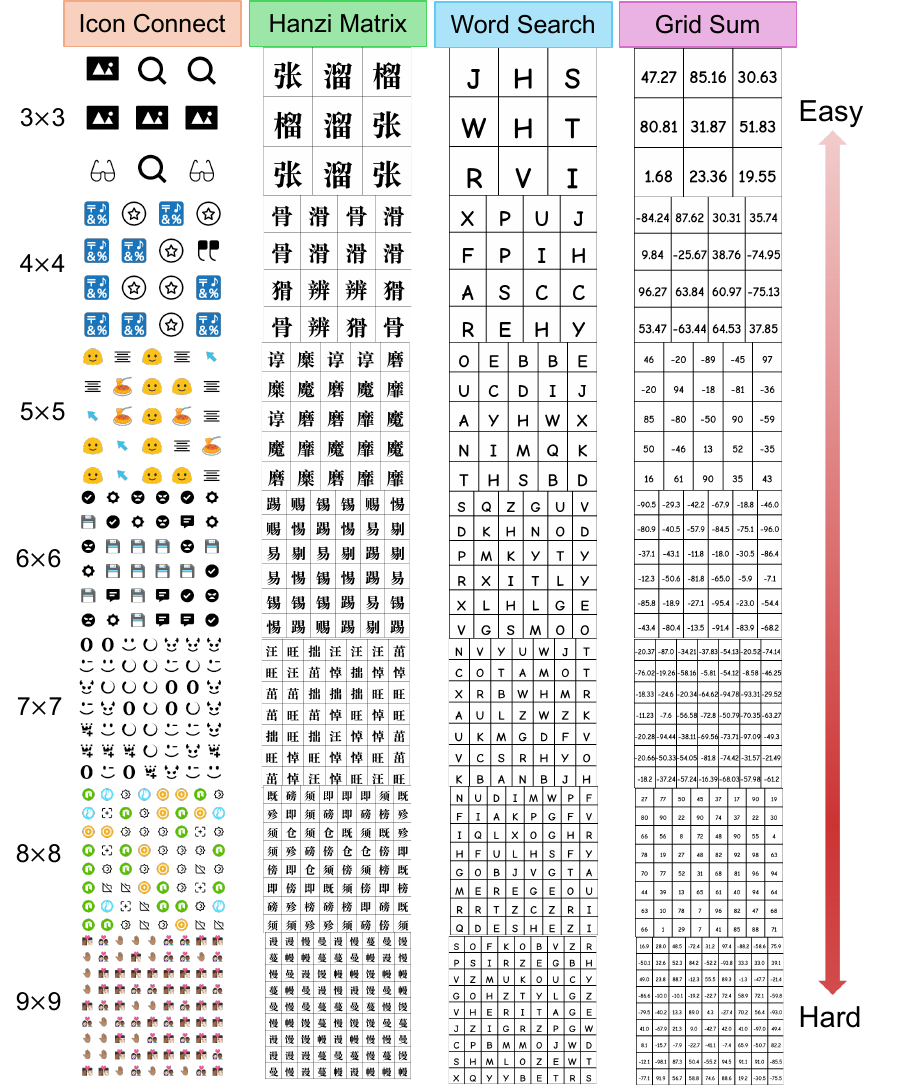}
  \caption{Overview of the difficulty level of the four puzzle tasks: Icon Connect, Hanzi Matrix, Word Search, and Grid Sum.}
  \label{difficult}
\end{figure}

\noindent\textbf{Icon Connect}. The goal of this puzzle is to determine the number of consecutive occurrences of a specified icon within an image grid. Images are generated at a fixed resolution of 512×512 pixels, with grid sizes varying between 3 and 9. Icons are sampled from \textit{Iconify}, an open-source repository containing over 200,000 vector icons. Sampling conditions include the number of distinct icon types and their color characteristics, where richly colored icons potentially increase difficulty for large multimodal models (LMMs). After sampling, icons are randomly positioned within a row or column of the grid, with a predefined number of consecutive repetitions, constrained by the grid size. These random placement parameters are recorded to generate corresponding question-answer pairs.

\noindent\textbf{Hanzi Matrix}. This puzzle focuses on counting distinct Chinese characters (hanzi) within an image grid. Images have a resolution of 512$\times$512 pixels, with grid sizes ranging from 3 to 9. Hanzi characters are selected from a collection comprising over 20,000 characters, including 500 groups of visually similar characters. Sampling prioritizes groups of visually similar characters; if the selected group does not fulfill the required quantity, additional characters are randomly sampled from non-similar sets. The number of sampled characters is recorded to produce accurate question-answer pairs.

\noindent\textbf{Word Search}. This puzzle requires identifying specified words embedded within a grid of letters. Images maintain a resolution of 512×512 pixels, with grid sizes from 3 to 9. Visual elements consist of the 26 letters of the English alphabet, with target words randomly chosen from a set of over 3,000 common words. The length of each selected word does not exceed the grid size, ensuring the word appears either horizontally or vertically. The designated word is randomly placed within a row or column, and remaining grid cells are filled with random letters. The exact positions of the words are recorded for generating corresponding question-answer pairs.

\noindent\textbf{Grid Sum}. The objective of this puzzle is to calculate the sum of all numerical values within a numeric image grid. Images are generated at a resolution of 512$\times$512 pixels, with grid sizes between 3 and 9. Numbers sampled for the grids range from -100 to 100 and are categorized based on their sign distribution (positive, negative, or mixed) and numerical precision (integer, one decimal, or two decimals). The sampled numbers are arranged into an $ N\times N$ grid, and the sum of all grid values is automatically calculated and recorded, forming the basis for the generated question-answer pairs.

\section{Difficult level of Puzzle Tasks}

\noindent\textbf{Jigsaw}. This puzzle involves identifying a missing image fragment (tile). A total of 960 prompts across 12 categories (80 per category) were generated using GPT-4, with representative prompts visualized in a word cloud (Fig.\ref{cloud}). Stable Diffusion 3.5 Medium was employed to produce a source pool of AI-generated images (AIGIs), generating two images per prompt, totaling 1,920 images. Images are available in three resolutions (256$\times$256, 512$\times$512, and 1024$\times$1024). After resizing, images are divided into $N \times N$ tiles using two segmentation approaches: regular (uniform partitions) and random (varying shapes and sizes). Four tiles are randomly selected as answer choices, one of which is designated as the correct answer, and a corresponding incomplete image missing the correct tile is generated. These configurations are recorded to generate question-answer pairs.

\noindent\textbf{Difference Hunt}. The objective of this puzzle is to detect subtle differences between pairs of similar images. The resource pool is identical to the Jigsaw puzzle, utilizing the same set of AI-generated images (AIGIs). Images are available at resolutions of 256$\times$256, 512$\times$512, and 1024$\times$1024 pixels. After resizing, images undergo randomized quality distortions, including underexposure, overexposure, noise, blur, and color inversion. Each distortion is applied to distinct, non-overlapping regions. The puzzle includes a configurable difficulty level from 1 (easiest) to 5 (hardest); higher difficulty levels feature smaller and less pronounced distortion areas. The exact number of distortion areas is recorded, enabling the generation of precise question-answer pairs.

\begin{figure}
  \centering
  \includegraphics[width=1\linewidth]{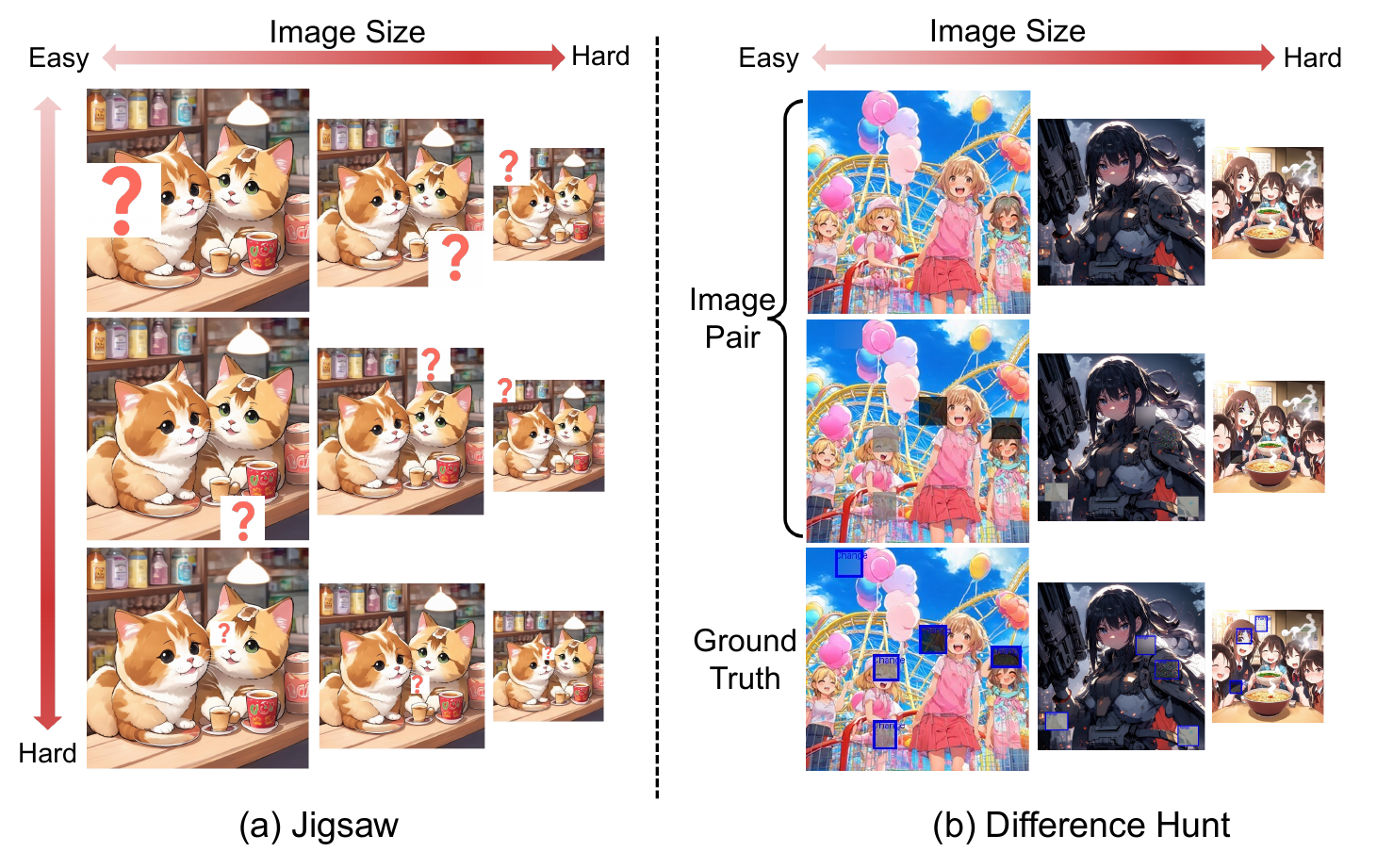}
  \caption{Difficulty level of two puzzle tasks. (a) Jigsaw. (b) Difference Hunt.}
  \label{Jigsaw_diff}
\end{figure}

When designing the puzzle tasks, we explicitly considered their difficulty levels. For grid-image-based puzzle tasks, including Icon Connect, Hanzi Matrix, Word Search, and Grid Sum, the difficulty level is closely associated with grid size, as illustrated in Fig.\ref{Jigsaw_diff} Given a fixed resolution of the puzzle images, an increase in grid size corresponds directly to smaller visual elements occupying each cell. This reduction in visual element size significantly challenges LMMs by demanding a finer-grained visual recognition capability. Thus, visual recognition represents a fundamental evaluation dimension for these four puzzle tasks. Furthermore, Word Search and Grid Sum tasks require not only accurate visual identification but also additional logical reasoning. Due to the distinct objectives inherent in each puzzle type, we adopt grid size as the principal criterion to define their difficulty levels.

In AI-generated-image-based (AIGI-based) puzzle tasks, specifically the Jigsaw and Difference Hunt tasks, we applied different difficulty criteria. For the Jigsaw puzzle, images are systematically divided into $N \times N$ tiles across three varying image resolutions. Lower resolutions coupled with higher tile counts (greater $N$ values) result in smaller-sized tiles, thereby significantly increasing the puzzle's difficulty. Conversely, for the Difference Hunt task, we introduced varying levels of distortion at three distinct image resolutions. In this case, lower image resolution combined with subtler distortion intensities creates higher difficulty, requiring LMMs to engage in more sophisticated and fine-grained contextual image understanding.

\end{document}